\documentclass[lettersize,journal]{IEEEtran}
\usepackage{amsmath,amsfonts}
\usepackage{algorithmic}
\usepackage{algorithm}
\usepackage{array}
\usepackage[caption=false,font=normalsize,labelfont=sf,textfont=sf]{subfig}
\usepackage{textcomp}
\usepackage{stfloats}
\usepackage{url}
\usepackage{verbatim}
\usepackage{graphicx}
\usepackage{cite}
\usepackage{xcolor} 
\usepackage{tcolorbox}
\usepackage{listings}
\usepackage{booktabs}
\usepackage{pgfplots}
\pgfplotsset{compat=1.18}
\usepackage{amssymb}
\usepackage{pifont}
\newcommand{\cmark}{\textcolor{green!60!black}{\ding{51}}}
\newcommand{\xmark}{\textcolor{red}{\ding{55}}}
\usepackage[hidelinks]{hyperref}
\tcbuselibrary{listings, skins, breakable}

\newtcolorbox{promptbox}[1][]{
    enhanced,
    breakable,
    colback=gray!5,
    colframe=gray!50,
    title=#1,
    left=6pt,
    right=6pt,
    top=6pt,
    bottom=6pt,
    boxrule=0.5pt,
    arc=2pt,
    fonttitle=\bfseries\footnotesize,
    fontupper=\footnotesize\ttfamily,
    before upper={\setlength{\parindent}{0pt}\setlength{\parskip}{6pt}}
}

\begin{document}

\title{Perception, Verdict, and Evolution: Hindsight-Driven Self-Refining Forensics Agent for AI-Generated Image Detection}

\author{
Yangjun~Wu,
Keyu~Yan,
Yu~Liu,
Jingren~Zhou,
Fei~Huang,
Rong~Zhang,
Zhou~Zhao,
and Fei~Wu%
\thanks{Yangjun Wu, Zhou Zhao, and Fei Wu are with Zhejiang University, Zhejiang, China (e-mail: 12460105@zju.edu.cn; zhaozhou@zju.edu.cn; wufei@zju.edu.cn).}%
\thanks{Yangjun Wu, Keyu Yan, Yu Liu, Jingren Zhou, Fei Huang and Rong Zhang are with Alibaba Group (e-mail: shuxin.wyj@alibaba-inc.com; yankeyu.yky@alibaba-inc.com; ly103369@alibaba-inc.com; jingren.zhou@alibaba-inc.com; feirhuang@gmail.com; stone.zhangr@alibaba-inc.com).}%
\thanks{Yangjun Wu is the first author and the corresponding author.}
}
        % <-this % stops a space

\markboth{Journal of \LaTeX\ Class Files,~Vol.~14, No.~8, August~2021}%
{Shell \MakeLowercase{\textit{et al.}}: A Sample Article Using IEEEtran.cls for IEEE Journals}


\maketitle

\begin{abstract}
The rapid advancement of generative models presents a significant challenge to existing deepfake detection methods, particularly given the widespread dissemination of highly realistic AI-generated images. Although Multimodal Large Language Models (MLLMs) show strong potential for this task, existing approaches suffer from two key limitations: insufficient sensitivity to fine-grained forensic artifacts and reliance on static synthetic supervision from frontier models, leading to limited flexibility and high-cost. 
% current methods rely heavily on fixed image-editing pipelines or synthetic data from frontier models,  leading to limited flexibility and high-cost. 
To address these issues, we propose \textit{ForeAgent}, an agentic forensics framework for AI-generated image detection with iterative self-evolution.
First, \textit{ForeAgent} adopts a Perception-Verdict architecture that aggregates multi-view cues spanning semantic, spatial, and frequency-domain features, and leverages an MLLM as a verdict module to fuse these signals for a logical-grounded verdict. 
Second, to enable continual self-improvement, we introduce a \textit{Hindsight-Driven Self-Refining} strategy following a \textit{Sampling-Reflection-Evolution} paradigm. The agent performs inference rollouts on training instances. Guided by ground-truth labels as hindsight, it reflects on failure cases and low-quality reasoning trajectories to regenerate higher-quality reasoning traces. These synthesized samples are then strictly filtered through a dual-expert quality gating module. \textit{ForeAgent} continuously evolves via fine-tuning on self-curated high-quality samples. Extensive experiments demonstrate that ForeAgent achieves state-of-the-art performance on the Chameleon benchmark, reaching 82.18\% accuracy (+16.41\% over AIDE), and achieves 93.3\% mean accuracy on AIGCDetectBenchmark across 16 generators. In addition, external evaluation shows that ForeAgent produces more consistent and causally grounded reasoning compared to GPT-5 and GPT-5-mini.
% Experimental results demonstrate that ForeAgent achieves SOTA performance on the Chameleon with an accuracy of 82.18\% (outperforming AIDE by 16.41\%) and marginally surpasses AIGI-homes  on the AIGCDetectBenchmark. Furthermore, quantitative evaluation of reasoning quality demonstrates that \textit{ForeAgent} achieves a higher reasoning score than both GPT-5 and GPT-5-mini.
\end{abstract}


\section{Introduction}
% 列举模型能力变强，图逼真，社交媒体上用户使用越来越多，检测任务变得越发重要
\IEEEPARstart{R}{ecent} advances in generative models for image synthesis have significantly improved the visual fidelity of synthetic images, narrowing the perceptual gap between authentic and AI-generated content. Representative paradigms include Generative Adversarial Networks (GANs) \cite{NIPS2014_f033ed80}, Diffusion Models \cite{10.5555/3495724.3496298}, and Autoregressive Generation Models \cite{li2024autoregressive,tian2024visual}. The widespread dissemination of such content on social media platforms has raised substantial concerns in digital forensics \cite{Huh_2018_ECCV} and copyright protection \cite{ren2024copyrightprotectiongenerativeai}. Consequently, reliable detection of AI-generated images has become a critical research problem for both academic study and real-world deployment.

% \IEEEPARstart{T}{he} remarkable progress in image generation technology has led to the synthesis of increasingly photorealistic images, blurring the boundary between authentic and artificial visual content. These methods involve Generative Adversarial Networks(GANs) \cite{NIPS2014_f033ed80}, Diffusion Models 
% \cite{10.5555/3495724.3496298}, and Autoregressive Generation Models \cite{li2024autoregressive,tian2024visual}. As these AI-generated images proliferate across social media platforms, they pose significant challenges to critical domains such as image forensics \cite{Huh_2018_ECCV} and copyright protection \cite{ren2024copyrightprotectiongenerativeai}. Consequently, the task of distinguishing AI-generated images from authentic ones has emerged as an urgent priority for both academic research and real-world applications.

\begin{figure}
    \centering
    \includegraphics[width=.5\textwidth]{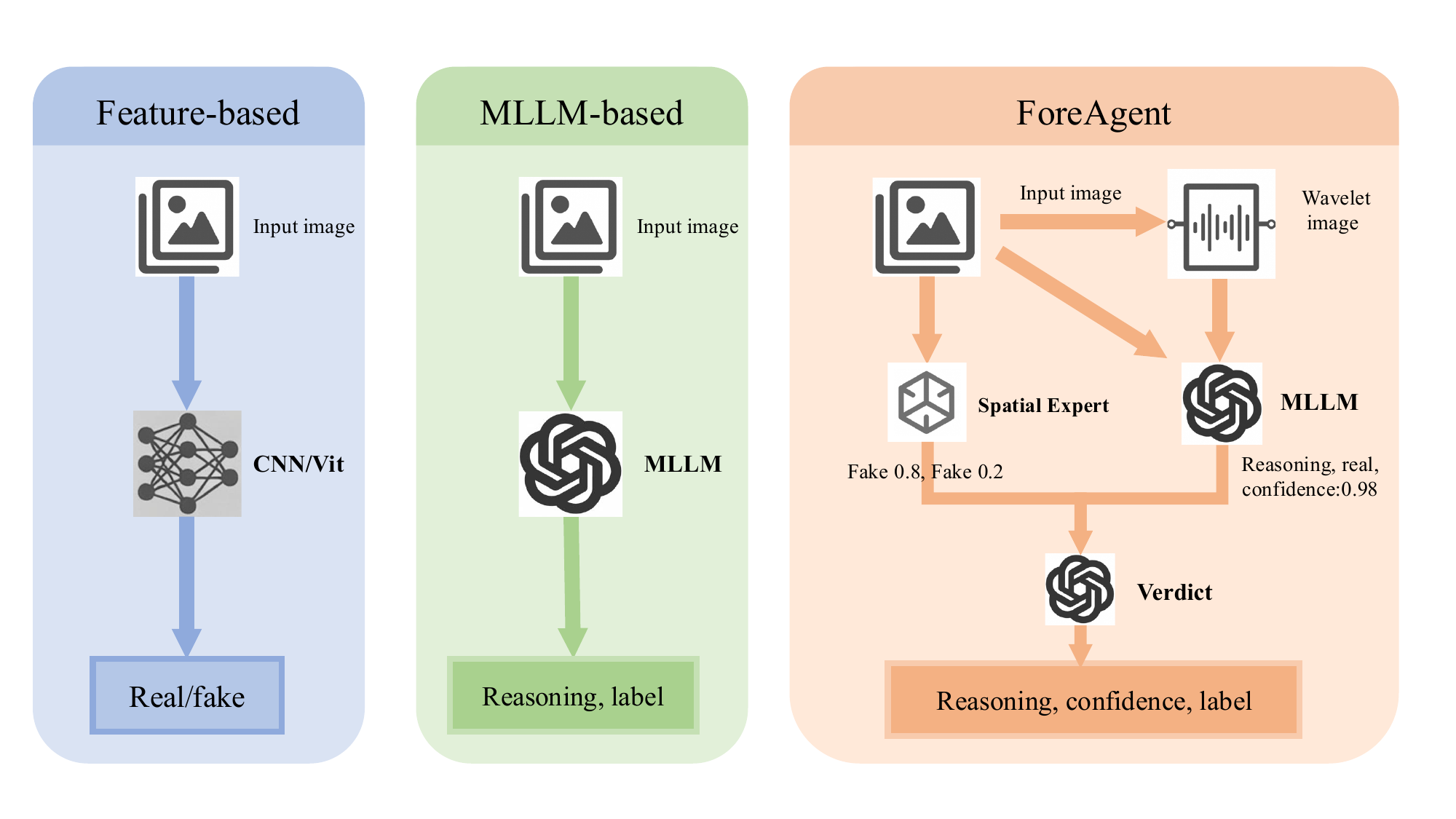}
    \caption{Comparison of our method with feature-based and MLLM-based methods. Here, \emph{reasoning} denotes the reasoning process. }
    \label{fig:comparison method} 
\end{figure}

% 列举现有方法
% Feature-based问题：
% ● 这些方法在捕获低级统计痕迹（信号层）方面表现卓越，但其本质是启发式的模式匹配。
% ● 缺乏高层语义推理能力，对图像内在的逻辑自洽性（如物理规律、常识逻辑）没有推理能力
% MLLMs-based问题：
% ● 依赖 固定模式的图像编辑 和 SOTA 多模态模型合成reasoning数据，推理能力高度依赖于外部 SOTA 模型（如 GPT-4o）生成的静态合成指令数据。
% ● 知识滞后性：无法应对最新生成器产生的独特错误
% ● 缺乏自我净化能力：无法通过反思失败来实现能力的迭代式演进
Existing AI-generated image detection methods primarily fall into two categories: \textbf{Feature-based} detectors and Multimodal Large Language Model \textbf{MLLMs-based} explainable methods. Early feature-based approaches \cite{8803661,wang2019cnngenerated,fredetect,karras2018progressive,effort,tan2023rethinking,Chameleon}, such as NPR \cite{tan2023rethinking} and AIDE \cite{Chameleon}, focus on extracting statistical artifacts. For instance, NPR captures up-sampling traces by analyzing neighboring pixel relationships, while AIDE extracts  frequency and semantic features to screen anomalies. However, these methods are typically pattern-matching schemes, lacking high-level semantic reasoning capabilities and cannot assess the intrinsic logical consistency of an image.
% treat detection as a binary classification task, lacking the logical reasoning required to explain why an image is fake. 
% 多模态大模型让可解释性的推理有了长足的进步, 相关论文还需要扩增，防止被攻击。
% 多模态大语言模型（Multimodal Large Language Models, MLLMs）涌现出的强大能力为可解释的AI生成图像检测带来了显著进展。例如，近期研究利用InternVL2.5-78B、Pixtral-124B等多个前沿多模态模型合成推理输出，并结合人工专家标注进行精化；FakeShield则借助精心构建的图像编辑流程，联合GPT-4o生成关于篡改区域、操纵内容及伪造痕迹的细粒度描述。然而，当前基于多模态大模型的方法仍存在两方面的关键局限：（1）尽管MLLMs展现出强大的通用视觉理解能力，但其对AI生成图像特有的细微伪影及局部异常的感知能力仍较为有限；（2）现有方法高度依赖GPT-4o、Pixtral-124B等闭源前沿模型，而对于基于错误分类样本的反思学习以及迭代自优化机制尚缺乏深入探索。
% The emergence of MLLMs has fostered explainable forensics . Methods like and 
The emergent capabilities of MLLMs have significantly advanced explainable AI-generated image detection \cite{SIDA,fakereasoning,yang2024researchabilityllmtamperdetection,zhang2024commonsensereasoningdeepfake,zhou2025aigi,kang2025legion}. For instance, AIGI-Holmes \cite{zhou2025aigi} leverages multiple SOTA multimodal models such as InternVL2.5-78B and Pixtral-124B \cite{agrawal2024pixtral12b} to synthesize reasoning outputs, which are subsequently refined through expert annotation. Similarly, FakeShield \cite{xu2024fakeshield}  exploits curated image editing pipelines in conjunction with GPT-4o \cite{openai2024gpt4ocard} to generate fine-grained descriptions of tampered regions, manipulation content, and forensic traces. Despite these advances, current MLLM-based approaches suffer from two critical limitations: 1) while MLLMs demonstrate strong general visual understanding, they exhibit limited sensitivity to subtle artifacts and fine-grained anomalies characteristic of AI-generated imagery. 2) Existing methods heavily dependent on externally provided, static synthetic instruction data produced by models such as GPT-4o or Pixtral-124B. leaving the potential of reflective learning on misclassified samples and iterative self-refinement largely unexplored. 
% This leads to a lag in knowledge: these systems struggle to handle novel or idiosyncratic failure modes introduced by the latest generation models.

% 我们的做法: ForeAgent 的定义是什么？
% 我们提出了一个Hindsight-Driven Self-Refining 的agentic框架，通过自我的反思迭代，ForeAgent实现了 real/fake辨别的准确率，且推理质量高于gpt5。 hindsight-driven self-refining
% To address these limitations, we propose \textbf{ForeAgent}, an agentic framework that incorporates multi-perspective visual cues to enable comprehensive forensic analysis. Crucially, the framework evolves through a hindsight-driven self-refinement mechanism, wherein it iteratively reflects on erroneous predictions or suboptimal reasoning quality, thereby progressively enhancing both detection capability and interpretability.
% 具体地，ForeAgent
To address these limitations, we propose \textbf{ForeAgent}, an agentic forensic framework for AI-generated image detection built upon an iterative evolutionary strategy. The framework integrates multi-perspective visual cues to enable comprehensive forensic analysis and evolves through a hindsight-driven self-refinement mechanism (guided by ground-truth labels), progressively improving both detection performance and interpretability. First, ForeAgent adopts a \textbf{Perception--Verdict} architecture to overcome the limitations of single-view detection. Unlike prior methods~\cite{zhou2025aigi,kang2025legion} that modify backbone architectures by introducing dual-branch feature extractors (e.g., CNNs combined with MLLMs) and training task-specific classification heads, ForeAgent preserves the MLLM in its original form and leverages it as a centralized reasoning engine without architectural modification. Specifically, ForeAgent constructs a comprehensive perceptual input consisting of three complementary dimensions as depicted in Figure~\ref{fig:comparison method}: (1) \emph{Semantic cues} derived from the original image to capture high-level visual inconsistencies; (2) \emph{Frequency-domain cues} obtained via \href{https://en.wikipedia.org/wiki/Wavelet_transform}{\textit{wavelet transforms}} to expose spectral anomalies that are often imperceptible to human inspection. These two modalities are jointly provided to the MLLM as dual-image inputs;  (3) \emph{Spatial cues} extracted by invoking an external NPR (Neighboring Pixel Relationships) detector as a callable tool, which outputs structured inference results rather than raw feature maps. The MLLM further serves as a \textit{verdict} module that integrates these multi-dimensional signals into a coherent and logically grounded decision, enabling holistic and evidence-aware forensic reasoning.

Second, to reduce reliance on external frontier MLLMs, image-editing or costly human annotations, we introduce a \textbf{Hindsight-Driven Self-Refining} strategy following a \textit{Sampling--Reflection--Evolution} paradigm. During training, the agent performs inference rollouts to identify hard cases characterized by incorrect predictions or low-quality reasoning traces (details in supplemental materials). Guided by ground-truth labels as hindsight signals, the agent reflects on these failure cases and regenerates improved reasoning trajectories. The synthesized samples are subsequently filtered through a dual-expert quality-gating module. High-quality reasoning samples are used for supervised fine-tuning, whereas samples with correct predictions but suboptimal reasoning are retained for label-only supervision to preserve foundational knowledge. Through this iterative process, failure cases are transformed into high-value training signals, enabling continuous self-evolution of ForeAgent.

Our contributions are summarized as follows:
\begin{itemize}
    \item[(1)] We propose ForeAgent, an agentic multi-view forensic framework that unifies semantic, spatial, and frequency-domain cues under a Perception–Verdict architecture without modifying the MLLM backbone.
    \item[(2)] We introduce a Hindsight-Driven Self-Refining paradigm that converts failure cases into high-value supervision signals through Sampling–Reflection–Evolution with dual-expert quality gating, enabling continual self-evolution.
    % We introduce a \textbf{Hindsight-Driven Self-Refining} mechanism that converts failure cases and low-quality outputs into high-value training signals via a Sampling--Reflection--Evolution paradigm and dual-expert quality gating, enabling self-evolution through fine-tuning.
    \item[(3)] Extensive experiments on AIGCDetectBenchmark and Chameleon demonstrate that ForeAgent achieves state-of-the-art performance and superior reasoning quality compared to GPT-5 and existing explainable detectors. The ForeAgent model will be be publicly available at \url{https://huggingface.co/Shimin/qwen3_vl_8b_foreagent}
    % ForeAgent achieves state-of-the-art performance on Chameleon with 82.18\% accuracy (+16.41\% over AIDE), while modestly exceeding AIGI-Homes on AIGCDetectBenchmark. Furthermore, quantitative assessment of reasoning quality reveals that ForeAgent outperforms both GPT-5 and GPT-5-mini.
    % ForeAgent achieves state-of-the-art performance on Chameleon (82.18\% accuracy, +16.41\% over AIDE) and competitive results on AIGCDetectBenchmark, while producing reasoning explanations that are qualitatively superior to those of GPT-5.
\end{itemize}

\begin{figure*}
    % [htbp] 是浮动位置选项，h:here, t:top, b:bottom, p:page % 包含 
    \centering
    \includegraphics[width=1.0\textwidth]{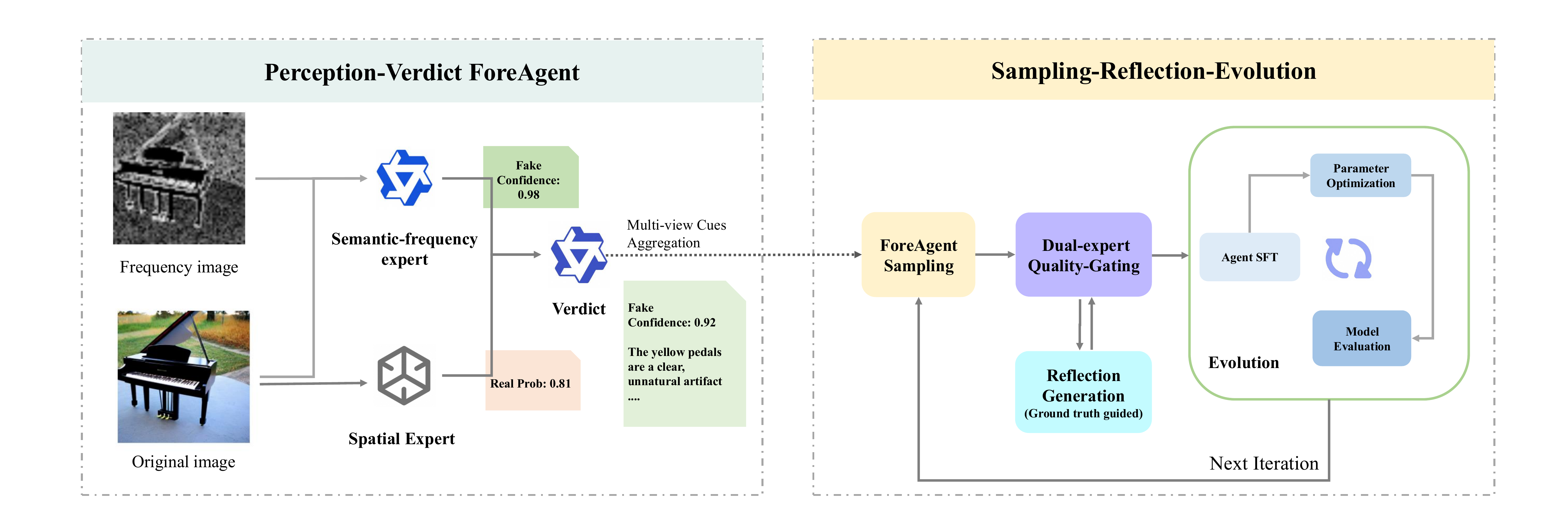}
    \caption{\textbf{Overview of ForeAgent.} ForeAgent combines a multi-view \textit{Perception--Verdict} architecture (left) with a \textit{Hindsight-Driven Self-Refining} loop (right). A semantic--frequency expert and a spatial expert extract complementary cues, which a verdict module fuses into a confidence-calibrated verdict with interpretable rationale. Through iterative \textit{Sampling--Reflection--Evolution}, the agent: (1) performs inference rollouts to collect predictions (\textit{Sampling}); (2) reflects on failure cases guided by ground-truth labels to regenerate improved reasoning traces (\textit{Reflection}); and (3) filters high-quality samples via dual-expert quality gating and fine-tunes itself for the next iteration (\textit{Evolution}). This closed-loop paradigm enables continual self-improvement in AI-generated image detection.}
    \label{fig:framework} % 方便交叉引用
\end{figure*}
\vspace{-2pt}

\section{METHODOLOGY}
% In this section, we present \textbf{ForeAgent}, an agentic forensics framework for discerning authentic and synthetic images. The framework consists of two core designs: (1) a \textit{Perception--Verdict} architecture that aggregates multi-view cues and leverages an MLLM-based verdict module for logical verdict generation, and (2) a \textit{Hindsight-Driven Self-Refining} mechanism following a \textit{Sampling--Reflection--Evolution} paradigm for iterative self-evolution. An overview of ForeAgent is illustrated in Figure~\ref{fig:framework}.
In this section, we present \textbf{ForeAgent}, an agentic forensics framework for discerning authentic and synthetic images. The framework includes two core designs: (1) a \textit{Perception--Verdict} architecture and (2) a \textit{Hindsight-Driven Self-Refining} mechanism. The overview is illustrated in Figure~\ref{fig:framework}.

%----------------------------------------------------------
\subsection{Multi-view Cues Perception--Verdict}
\label{sec:perception_verdict}
%----------------------------------------------------------
% 需要强调和mllm利用npr的不同，我们不希望改造mllm的架构
To overcome the limitations of single-view detection, ForeAgent adopts a \textit{Perception--Verdict} architecture that aggregates multi-view cues spanning semantic, frequency-domain, and spatial dimensions. Rather than simply concatenating heterogeneous signals or training additional fusion modules,we employ an MLLM (Qwen3-VL-8B) as the central \textit{reasoning agent}. The agent actively perceives semantic and frequency-domain cues through dual-image input, while invoking an external spatial expert (NPR detector \cite{tan2023rethinking}) as a callable tool to obtain spatial-domain evidence. A dedicated \textit{Verdict} module then synthesizes these multi-dimensional findings, enabling the agent to produce a coherent, logically grounded, and fully interpretable detection conclusion.

\subsubsection{Multi-View Feature perception} The input feature comprises: 1) Semantic cue (original image). This captures high-level visual inconsistencies and semantic artifacts that may indicate AI generation, such as unnatural object compositions, implausible lighting, or anomalous facial features. 2) Frequency-Domain cue (wavelet transform). We employ wavelet transformation to extract diagonal detail maps with fixed size \textit{(256,256)}, which expose frequency-domain anomalies that are often imperceptible in the pixel domain. 
% These cues are particularly effective for detecting periodic patterns and spectral inconsistencies introduced by generative models.
3) Spatial Cue (Optional). We employ the NPR detector \cite{tan2023rethinking} to capture structural traces left by up-sampling operations inherent in generative pipelines. In this study, we direct use this expert without tuning for simplicity. 
% NPR features encode local pixel correlations that differ systematically between authentic and synthesized images.

The formulation of the standard detection task is that we have the image $x$ and its related label $y$:
\begin{equation}
     y = \Theta_{model}(x) \ \  y\in \{\text{real},\, \text{fake}\}
    \label{eqa:1}
\end{equation}
$\Theta$ refers to the learnable parameters. In our multi-view setting, ForeAgent feeds the original image $\mathbf{x}$ together with its frequency-domain counterpart  $\mathbf{x}_{freq}$ as a joint (dual-image) input to the MLLM for semantic and spectral perception. Meanwhile, ForeAgent sends the original image $\mathbf{x}$ to the NPR expert and obtains its inference prediction. Thus, we revise the formulation as follows:
\begin{equation}
    \mathbf{y}_{1} = \mathcal{F}_{\text{MLLM}}(\mathbf{x}, \mathbf{x}_{freq})
\end{equation}
\begin{equation}
    \mathbf{y}_{2} = \mathcal{F}_{\text{NPR}}(\mathbf{x})
\end{equation}
where $\mathbf{y}_{1} = \{ C_{conf}, R_{reasoning},P_{pred}\}$ comprises the binary prediction $P_{pred} \in \{\text{real}, \text{fake}\}$, the confidence score $C_{conf} \in [0, 1]$, and the reasoning explanation $R_{reasoning}$ that provides interpretable justification for the decision. Simultaneously, the original image $\mathbf{x}$ is processed by spatial expert to extract low-level artifacts. $\mathbf{y}_{2} = \{Pr_{real}, Pr_{fake}\}$ represents the probability distribution over real and fake classes.

This design enables ForeAgent to capture high-level semantic inconsistencies as well as frequency-domain artifacts via the MLLM , and low-level statistical fingerprints by the NPR expert, providing complementary evidence for discerning authentic and synthetic images.

\subsubsection{MLLM-Based Verdict} The multi-view \emph{evidence} obtained from different perspectives is fed into an MLLM-based \textit{verdict module} (Qwen3-VL-8B as the backbone), which serves as the reasoning core of ForeAgent. Unlike conventional classifiers that directly output binary predictions from learned feature embeddings, our verdict module performs explicit logical arbitration over \emph{inference-level signals} (i.e., perceptual findings and expert predictions) to render the final verdict. Specifically, the verdict module is prompted to:
\begin{itemize}
    \item Inspect each view independently and summarize potential forgery indicators.
    \item Integrate and cross-check evidence across views to resolve conflicts and ambiguities.
    \item Produce a coherent rationale that culminates in a \texttt{Real} or \texttt{Fake} verdict with an associated confidence score.
\end{itemize}
The verdict module produces the final decision as:
\begin{equation}
    \hat{y} = \mathcal{F}_{\text{MLLM}}\big(p, \mathbf{x}, \mathbf{x}_{freq}, \mathbf{y}_{1}, \mathbf{y}_{2}\big)
\end{equation}
Formally, $p$ is the system prompt (see supplemental materials verdict prompt), $\hat{y}$ contains reasoning, prediction label $v_{pred}$ and confidence. This design ensures holistic and evidence-based judgment, while providing interpretable reasoning traces aligned with the multi-view forensic evidence.

% The extracted multi-view features are provided to an MLLM-based verdict module (Qwen3-VL-8B), which serves as the reasoning core of ForeAgent. Unlike conventional classifiers that directly output binary predictions, the verdict module performs logical arbitration over the multi-dimensional signals to render a final verdict. Specifically, the verdict module is prompted to:
% \begin{itemize}
%     \item Analyze each view independently and identify potential forgery indicators;
%     \item Integrate evidence across views to resolve conflicts and ambiguities;
%     \item Generate a coherent reasoning that culminates in a \texttt{Real} or \texttt{Fake} verdict with an associated confidence score.
% \end{itemize}
% The formulation is: 
% \begin{equation}
%     \mathbf{y}_{2} = \mathcal{F}_{\text{MLLM}}(\mathbf{x}, \mathbf{x}_{freq},\mathbf{y}_{1\_1}, \mathbf{y}_{1\_2, p})
% \end{equation}
% This design ensures a holistic and evidence-based judgment and interpretable reasoning explanations.

%----------------------------------------------------------
\subsection{Hindsight-Driven Self-Refining}
\label{sec:hindsight}
%----------------------------------------------------------

To reduce the reliance on external frontier MLLMs as well as costly human annotations, we introduce a Hindsight-Driven Self-Refining strategy. This mechanism allows ForeAgent to sample instances in the training set and perform targeted reflection on failure cases and low-quality outputs, thereby synthesizing higher-quality reasoning traces for iterative fine-tuning. The overall procedure follows a \textit{Sampling--Reflection--Evolution} paradigm and the workflow of paradigm described in algorithm \ref{alg:workflow}.
% \begin{algorithm}
% \caption{The workflow of Hindsight-Driven Self-Refining.}
% \begin{algorithmic}
% \STATE 
% \STATE {\textsc{Sampling}}
% \STATE \hspace{0.5cm}$ y_2 = ForeAgent(x) \ x \subset \mathbf{D^{i}}  $
% \STATE \hspace{0.5cm}$ categorize(y_2, y) \ in \ [correct, incorrect] $
% \STATE \hspace{0.5cm}$ Quality = DEG(y_2, y) \ in \ [Low, High] $
% \STATE 
% \STATE {\textsc{Relefction}}
% \STATE \hspace{0.5cm}$ new_y = Relefction(y_2, y) $
% \STATE \hspace{0.5cm}$ Quality = DEG(new_y, y) \ in \ [Low, High] $
% \STATE \hspace{0.5cm}$  build \ training \ data $
% \STATE 
% \STATE {\textsc{Training}}$(\mathbf{X} )$
% \STATE \hspace{0.5cm}$ update \ \theta $
% \STATE \hspace{0.5cm}$ evaluation $
% \end{algorithmic}
% \label{alg1}
% \end{algorithm}

\begin{algorithm}[t]
\small
\caption{Iterative Hindsight-Driven Self-Refining}
\begin{algorithmic}[1]
\REQUIRE Training set $\mathcal{D}$, Test set $\mathcal{D}_{test}$, Initial model $\mathcal{M}_{\theta_0}$, Maximum iterations $T$, Data splits $N$
\ENSURE Refined model $\mathcal{M}_{\theta^*}$
\STATE Initialize best score $\text{S}_{best} \leftarrow 0$, best model $\theta^* \leftarrow \theta_0$
\STATE Partition $\mathcal{D}$ into $N$ subsets: $\{\mathcal{D}_1, \mathcal{D}_2, \ldots, \mathcal{D}_N\}$
\STATE

\FOR{$t = 1$ \TO $T$}
    \STATE Select subset $\mathcal{D}_t \leftarrow \mathcal{D}_{(t-1)} $
    \STATE Initialize refined sets: $\mathcal{D}_{high} \leftarrow \emptyset$, $\mathcal{D}_{label} \leftarrow \emptyset$
    \STATE
    
    \STATE \textbf{Phase 1: Sampling and Quality Assessment}
        \STATE Prediction: $\hat{y} \leftarrow \pi_{\theta_{t-1}}(x)$
        \STATE Prediction label: $v_{pred} \ from \ \hat{y} $, $v_{gt}$
        \IF{$v_{pred} = v_{gt}$}
            \STATE $C_x= (x,\mathbf{x}_{freq},\mathbf{y}_{1},\mathbf{y}_{2})$
            \STATE Evaluate quality: $q \leftarrow \text{DEG}(C_x, \hat{y}, v_{gt}) $
            \IF{$q \ is \ \text{High}$}
                \STATE $\mathcal{D}_{high} \leftarrow \mathcal{D}_{high} \cup \{(x, \hat{y})\}$
            \ELSE
                \STATE Trigger reflection
            \ENDIF
        \ELSE
            \STATE Trigger reflection
        \ENDIF
    \STATE
        
    \STATE \textbf{Phase 2: Reflection with Dual-Expert Validation}
        \IF{Reflection triggered}
            \STATE Generate reflection candidates via ForeAgent:
            \STATE \quad $tr_1 \leftarrow \text{Reflect}(x,x_{freq},y_1, v_{gt})$ for perception $y_1$
            \STATE \quad $tr_2 \leftarrow \text{Reflect}(\mathcal{C}_x, \hat{y}, v_{gt})$ for verdict $\hat{y}$
            \STATE Evaluate candidates via DEG:
            \STATE \quad $q_1 \leftarrow \text{DEG}(x, tr_1, v_{gt})$
            \STATE \quad $q_2\leftarrow \text{DEG}(\mathcal{C}_x,tr_2, v_{gt})$
            \STATE $q=max(q_1, q_2)$
            \STATE Appended to $\mathcal{D}_{high}$ if $q$ is High else $\mathcal{D}_{label}$
            % \STATE Appended to $\mathcal{D}_{label}$
            % \STATE $\mathcal{D}_{high} \leftarrow \mathcal{D}_{high} \cup \{(x, tr_1 \ or \ \mathcal{C}_x, tr_2)\}$ when $q$ is High
            % \STATE $\mathcal{D}_{label} \leftarrow \mathcal{D}_{label} \cup \{(x, v_{gt})\}$ when $q$ is not High
        \ENDIF
    \STATE
    
    \STATE \textbf{Phase 3: Model Evolution and Evaluation}
    \STATE $\theta_t \leftarrow \text{FineTune}(\theta_{t-1}, \mathcal{D}_{high} \cup \mathcal{D}_{label})$
    \STATE $\text{S}_t \leftarrow \text{Evaluate}(\mathcal{M}_{\theta_t}, \mathcal{D}_{test})$
    \IF{$\text{S}_t > \text{S}_{best}$}
        \STATE $\text{S}_{best} \leftarrow \text{S}_t$
        \STATE $\theta^* \leftarrow \theta_t$
    \ELSE
        \STATE $\theta_t \leftarrow \theta^*$
    \ENDIF
\ENDFOR
\STATE
\RETURN $\mathcal{M}_{\theta^*}$
\end{algorithmic}
\label{alg:workflow}
\end{algorithm}

\subsubsection{\textbf{Sampling}}
We partition the training set into $N$ disjoint shards $\{\mathcal{D}_{i}\}_{i=1}^{N}$ and perform iterative self-refinement over these shards. At iteration $t$, ForeAgent runs inference rollouts $\hat{y}$ with the current agent $\pi_{\theta_{t-1}}$ on the selected shard $\mathcal{D}_t$ and categorizes each trajectory into three groups: \textit{(1) Incorrect Predictions}, where the predicted label $v_{pred}$ differs from the ground-truth label $v_{\mathrm{gt}}$. \textit{(2) Low-Quality Reasoning}, where the prediction is correct ($v_{pred}=v_{\mathrm{gt}}$) but the generated rationale $\hat{R}$ is vague, generic, or logically flawed. \textit{(3) High-Quality Reasoning}, where both the prediction  is correct and the rationale is deemed evidence-grounded. The first two groups are treated as \emph{hard cases} and are prioritized for the subsequent reflection stage, while the high-quality group can be retained as stable supervision.

\subsubsection{\textbf{Dual-Expert Quality Gating (DEG)}}
\label{sec:quality_gating}
To ensure that the synthesized reasoning traces are both logically rigorous and evidentially grounded, we propose a multi-level dual-expert quality gating module. The module follows a \textbf{``lenient entry, strict exit''} principle: a sample is \emph{admitted} for reflection as long as \emph{either} expert identifies any quality deficiency—specifically, when any of the following conditions is detected: \texttt{logic consistent = false}, \texttt{evidence sufficient = false}, \texttt{causal clear = false}, or \texttt{quality score} $\leq 4$. A sample is incorporated into the high-quality dataset $\mathcal{D}_{high}$ \emph{if and only if} both experts agree that all quality criteria are satisfied.
% a sample is \emph{admitted} for reflection as long as \emph{either} expert judges its reasoning quality $q$ to be insufficient (i.e., a low-quality flag from one model is enough to trigger reflection). A sample is incorporated into the high-quality dataset $\mathcal{D}_{high}$ \emph{if and only if} both experts agree that the reasoning quality meets the high standard.

\noindent\textbf{Why Dual Experts?} We employ two independent MLLMs as judges: ForeAgent (Qwen3-VL-8B) and \href{https://modelstudio.console.alibabacloud.com/?spm=a3c0i.28768018.4869165850.1.689d5c10DejqTA&tab=doc#/doc/?type=model&url=2840914_2&modelId=qwen3-vl-plus}{\textit{Qwen3-VL-Plus}}. The rationale for adopting a dual-expert framework is empirically validated in our ablation study (Section~\ref{ablation:quality_gating}), which demonstrates that the evaluation of dual-experts effectively mitigates bias of a single-model and significantly improves the reliability of the evaluation compared to approaches of a single-judge. 

% \noindent\textbf{Sampling.} We actively split where the current agent exhibits suboptimal performance. In each iteration $t$, we deploy the agent $\pi_{\theta_t}$ to perform inference rollouts on a training subset and identify two types of problematic cases: \textit{Incorrect Predictions.} Cases where the agent's predicted label $\hat{y}$ deviates from the ground-truth label $y_{\text{gt}}$ (\emph{e.g.}, misclassifying an AI-generated image as real). \textit{Low-Quality Reasoning.} 

% \noindent\textbf{Reflection.} The sampled failure cases trigger a \textbf{Hindsight Reflection} process. Guided by ground-truth labels as hindsight, the agent is prompted to reflect on its errors and regenerate higher-quality reasoning traces: \textbf{For Incorrect Predictions.} The agent is prompted to re-examine the image given the ground-truth label $y_{\text{gt}}$ and generate a corrected reasoning path $R_{\text{corrected}}$ that logically leads to the correct verdict. \textbf{For Low-Quality Reasoning.} The agent is prompted to refine the existing weak rationale into a detailed, logically sound, and persuasive explanation $R_{\text{refined}}$.

% 怎么进行反思的？
% 质量和流程需要再明确一些，否则reviewer看不懂
\subsubsection{\textbf{Reflection}} 
The first two groups (\textit{Incorrect Predictions} and \textit{Low-Quality Reasoning}) are fed into a \textbf{Hindsight Reflection} process. Using the ground-truth label $v_{\mathrm{gt}}$ as hindsight, ForeAgent re-examines each original image $x$ along with its frequency-domain counterpart  $\mathbf{x}_{freq}$ and regenerates two candidate corresponding reasoning traces $tr_1$ and $tr_2$ for perception (the system prompt can be found in Supplemental materials.): \textit{(1)} For samples with incorrect predictions, the agent generates the corrected reasoning path $tr_{\mathrm{corr}}$ that leads to the correct verdict. \textit{(2)} For samples with correct predictions but weak rationales, the agent refines the original output into an improved reasoning trace $tr_{\mathrm{ref}}$ with clearer logic and more concrete evidence. Both candidate traces are evaluated by the Dual-Expert Gating (DEG) module, yielding quality scores $q_1$ and $q_2$. We select the candidate with the higher score, denoted as $q^* = \max(q_1, q_2); tr^* =  tr_{q^*}$, and check whether its quality meets a predefined threshold $\tau_t$. The candidate is appended to $\mathcal{D}_{high}$ only if $q^* \geq \tau_t$ and both experts agree on high quality; otherwise, it is degraded to label-only supervision in $\mathcal{D}_{label}$.
% % The regenerated traces from both groups are subsequently passed to the quality-gating module for verification and ranking. This strict filtering prevents the propagation of flawed reasoning into subsequent training iterations.\noindent\textbf{Quality Assessment and Selection.} 

To accommodate evolving learning dynamics, we adopt \textbf{adaptive quality thresholds} across iterations: thresholds are set relatively low in early stages to encourage exploration and capture diverse failure patterns, then progressively tightened in later iterations to ensure only high-quality reasoning is incorporated.

\subsubsection{Self-Improvement Evolution via Fine-Tuning.}
The synthesized reasoning samples undergo rigorous quality verification (described in Section \ref{sec:quality_gating}) before being incorporated into the training set. The curated samples are classified into two types: (1) High-quality samples $D_{high}$, which pass all verification criteria and are used with full reasoning supervision. (2) Label-only samples $D_{label}$, which have correct labels but suboptimal reasoning quality and are degraded to label-only learning tasks.

These samples are then used to fine-tune the agent, enabling iterative self-evolution. After each training iteration, we evaluate the updated model $\theta_t$ on the test set $\mathcal{D}_{test}$. The model parameters are updated to the new checkpoint \emph{if and only if} the performance $S_t$ surpasses the previous best result $S_{best}$; otherwise, the model reverts to the best-performing checkpoint from earlier iterations. This performance-gated update strategy ensures monotonic improvement and prevents model degradation. Through this process, ForeAgent effectively converts failure cases into high-value training signals, progressively sharpening its detection capability and reasoning quality across successive refinement cycles.
% \subsubsection{Evolution: Self-Improvement via Fine-Tuning.} 
% The synthesized reasoning samples undergo rigorous quality verification (described in Section \ref{sec:quality_gating}) before being incorporated into the training set. The curated samples can be classified in two types: (1) High-quality samples, which pass all verification criteria with full reasoning supervision. (2) Label-only samples, which with correct labels but suboptimal reasoning quality are degraded to label-only learning tasks. 

% These samples are then used to fine-tune the agent, enabling iterative self-evolution. This process effectively converts failure cases into high-value training signals, allowing ForeAgent to progressively sharpen its detection capability and reasoning quality across successive refinement cycles.

%----------------------------------------------------------
\subsection{Training Pipeline}
\label{sec:training}
%----------------------------------------------------------

The overall training pipeline of ForeAgent consists of two phases:

% \paragraph{Phase 1: Foundational Pre-Training.}  
% We initialize the agent via supervised pre-training on a large-scale balanced dataset $\mathcal{D}_{\text{pre}}$ containing authentic and AI-generated images. This phase establishes fundamental discriminative capabilities.
\paragraph{Phase 1: Foundational Pre-Training.}
We initialize the agent via supervised pre-training on a balanced dataset $\mathcal{D}_{\mathrm{pre}}$ containing authentic and AI-generated images. In this phase, each sample is annotated with a binary label (\texttt{real} or \texttt{fake}), and the model is trained with cross-entropy loss. This stage establishes fundamental discriminative capabilities without reasoning supervision.

% \paragraph{Phase 2: Iterative Self-Refinement.}  
% The agent undergoes multiple iterations of the Sampling--Reflection--Evolution loop. In each iteration, the agent identifies its failure cases, reflects with hindsight guidance, generates improved reasoning traces, and fine-tunes on the quality-gated samples. The iterative process continues until convergence or target performance metrics are achieved.

% Through this hindsight-driven self-refining mechanism, ForeAgent achieves continuous self-evolution, progressively enhancing both detection accuracy and interpretability without requiring additional human annotations.

\paragraph{Phase 2: Iterative Self-Refinement.}
The agent undergoes multiple iterations of the \textbf{Sampling--Reflection--Evolution} loop. In each iteration $t$, ForeAgent first samples and reflects on its failure cases to generate improved reasoning traces. The quality-gated samples are then used to LoRA fine-tune the agent, after which the adapter weights are merged into the base model to obtain the updated agent $\pi_{\theta_{t+1}}$.

We evaluate the updated model on the test set. If the performance improves compared to the previous iteration, the new parameters are retained and the loop proceeds to the next iteration. Otherwise, we revert to the previous checkpoint $\pi_{\theta_t}$ and continue. The iterative process terminates when the maximum number of iterations $T_{\max}$ (Default $5$) is reached.

Through this hindsight-driven self-refining mechanism, ForeAgent achieves continuous self-evolution, progressively improving both detection accuracy and interpretability without requiring additional human annotations or reliance on proprietary frontier models.
% Through this hindsight-driven self-refining mechanism, ForeAgent achieves continuous self-evolution, progressively enhancing both detection accuracy and interpretability without requiring human annotations or frontier model such as GPT5.

\section{Experiments}

\begin{table*}[t]
\centering
\caption{Results on AIGCDetectBenchmark. We report per-generator accuracy (\%) and the mean accuracy across all generators.}
\label{tab:aigcdetectbenchmark}
\resizebox{\textwidth}{!}{%
\begin{tabular}{lccccccccccccccccc}
\toprule
\textbf{Model} & \textbf{ProGAN} & \textbf{StyleGAN} & \textbf{BigGAN} & \textbf{CycleGAN} & \textbf{StarGAN} & \textbf{GauGAN} & \textbf{StyleGAN2} & \textbf{WFIR} & \textbf{ADM} & \textbf{GLIDE} & \textbf{Midjourney} & \textbf{SD1.4} & \textbf{SD1.5} & \textbf{VQDM} & \textbf{Wukong} & \textbf{DALLE2} & \textbf{Mean} \\
\midrule
CNNSpot      & 100.0 & 90.2 & 71.2 & 87.6 & 94.6 & 81.4 & 86.9 & 91.7 & 60.4 & 58.1 & 51.4 & 50.6 & 50.5 & 56.5 & 51.0 & 50.5 & 70.8 \\
FreDet       & 99.4  & 78.0 & 82.0 & 78.8 & 94.6 & 80.6 & 66.2 & 50.8 & 63.4 & 54.1 & 45.9 & 38.8 & 39.2 & 77.8 & 40.3 & 34.7 & 64.0 \\
Fusing       & 100.0 & 85.2 & 77.4 & 87.0 & 97.0 & 77.0 & 83.3 & 66.8 & 49.0 & 57.2 & 52.2 & 51.0 & 51.4 & 55.1 & 51.7 & 52.8 & 68.4 \\
LNP          & 99.7  & 91.8 & 77.8 & 84.1 & 99.9 & 75.4 & 94.6 & 70.9 & 84.7 & 80.5 & 65.6 & 85.6 & 85.7 & 74.5 & 82.1 & 88.8 & 83.8 \\
LGrad        & 99.8  & 91.1 & 85.6 & 86.9 & 99.3 & 78.5 & 85.3 & 55.7 & 67.2 & 66.1 & 65.4 & 63.0 & 63.7 & 73.0 & 59.6 & 65.5 & 75.3 \\
UnivFD       & 99.8  & 84.9 & 95.1 & 98.3 & 95.8 & 99.5 & 75.0 & 86.9 & 66.9 & 62.5 & 56.1 & 63.7 & 63.5 & 85.3 & 70.9 & 50.8 & 78.4 \\
DIRE-G       & 95.2  & 83.0 & 70.1 & 74.2 & 95.5 & 67.8 & 75.3 & 58.1 & 75.8 & 71.8 & 58.0 & 49.7 & 49.8 & 53.7 & 54.5 & 66.5 & 68.7 \\
DIRE-D       & 52.8  & 51.3 & 49.7 & 49.6 & 46.7 & 51.2 & 51.7 & 53.3 & 98.3 & 92.4 & 89.5 & 91.2 & 91.6 & 91.9 & 90.9 & 92.5 & 71.5 \\
PatchCraft   & 100.0 & 92.8 & 95.8 & 70.2 & 100.0 & 71.6 & 89.6 & 85.8 & 82.2 & 83.8 & 90.1 & 95.4 & 95.3 & 88.9 & 91.1 & 96.6 & 89.3 \\
NPR          & 99.8  & 97.7 & 84.4 & 96.1 & 99.4 & 82.5 & 98.4 & 65.8 & 69.7 & 78.4 & 77.9 & 78.6 & 78.9 & 78.1 & 76.1 & 64.9 & 82.9 \\
AIDE         & 100.0 & 99.6 & 84.0 & 98.5 & 99.9 & 73.3 & 98.0 & 94.2 & 93.4 & 95.1 & 77.2 & 93.0 & 92.9 & 95.2 & 93.6 & 96.6 & 92.8 \\
\midrule
Qwen3-VL-8B  & 72.1  & 51.0 & 53.4 & 61.0 & 53.6 & 58.5 & 50.1 & 50.0 & 52.3 & 55.8 & 51.7 & 52.4 & 52.5 & 54.6 & 57.2 & 51.9 & 54.9 \\
Qwen3-VL-32B & 79.4  & 52.6 & 57.7 & 73.4 & 75.5 & 71.7 & 51.0 & 54.0 & 56.9 & 62.9 & 55.8 & 55.1 & 55.1 & 59.4 & 62.9 & 52.8 & 61.0 \\
AIGI-Holmes  & 100.0 & 98.4 & 94.5 & 97.0 & 100.0 & 95.2 & 98.9 & 95.7 & 88.4 & 91.5 & 81.6 & 91.3 & 91.4 & 90.9 & 89.5 & 85.3 & 93.2 \\
\midrule
\textbf{ForeAgent} & 99.9 & 98.2 & 83.1 & 96.1 & 99.9 & 80.5 & 99.7 & 63.4 & 91.5 & 97.0 & 96.0 & 98.3 & 98.2 & 95.6 & 97.1 & 98.9 & \textbf{93.3} \\
\bottomrule
\end{tabular}%
}
\end{table*}

% Fakeshield \cite{xu2025fakeshield},
\subsection{Experimental Settings} 
\subsubsection{Detector baselines.} For a fair comparison, we perform comparisons of our approach with existing state-of-the-art featured-based detectors, including CNNSpot \cite{wang2019cnngenerated}, FreDet \cite{fredetect}, Fusing\cite{Fusing}, LNP \cite{LNP}, LGrad \cite{LGrad}, Dire \cite{dire}, UnivFD \cite{UnivFD}, PatchCraft \cite{PatchCraft}, NPR \cite{tan2023rethinking}, AIDE \cite{Chameleon} and Effort \cite{effort}. Meanwhile, we also choose popular open-sourced MLLMs, comprising AIGI-Holmes \cite{zhou2025aigi},\ Qwen3-VL-8B-Instruct, Qwen3-VL-32B-Instruct \cite{qwen3vltechnicalreport}, \href{https://ai.meta.com/blog/llama-3-2-connect-2024-vision-edge-mobile-devices/}{Llama-3.2-11B-Vision-Instruct}, and Proprietary Models \href{https://openai.com/index/gpt-5-system-card/}{GPT5-mini and GPT5}.

% train/test set (mentioned in Section \ref{data_construction}) 
\subsubsection{Datasets.}
We conduct experiments on two benchmarks under different training set to ensure fair comparison with prior work. First, following the standard setting adopted in previous studies ~\cite{wang2019cnngenerated,LNP,tan2023rethinking,dire, UnivFD}, all methods are trained exclusively on four object categories (\textit{car}, \textit{cat}, \textit{chair}, \textit{horse}) generated by ProGAN \cite{karras2018progressive} and evaluated on AIGCDetectBenchmark \cite{PatchCraft}, which spans 16 diverse generators. This setting assesses cross-generator generalization under a unified training regime. Second, we randomly sample both 60,000 real/fake images from the Genimage \cite{GenImage} training split to perform foundational pre-training. Subsequently, we apply the proposed \textit{Hindsight-Driven Self-Refining} strategy on an additional 100,000 training samples, enabling iterative self-evolution through the \textit{Sampling--Reflection--Evolution} loop. Finally, ForeAgent is evaluated on the Chameleon set \cite{Chameleon} to verify detection performance.

% 参数介绍
% \paragraph{Implementation details.} We employ Qwen3-VL-8B-Instruct as our agent backbone. During the foundation LORA pre-training, we set the learning rate, LoRA rank, LoRA alpha optimizer, and torch.dtype as 5e-6, 64, 128, AdamW, and bfloat16. As for self-evolution stage, num generations of candidates and learning rate, training epoch are set to 2 and 1e-6, 1. The ForeAgent is trained on 8 NVIDIA A100 GPUs for 2 epoch  and 1 epoch during the self-evolution stage, respectively. Reference to existing AI-generated detection methods, we report classification accuracy for real, fake, and average precision in our experiments.
\subsubsection{Implementation Details.} 
We adopt Qwen3-VL-8B-Instruct as the backbone without architectural modificatio for our agent. During the foundation LoRA pre-training phase, we utilize the AdamW optimizer with a learning rate of $5 \times 10^{-6}$, setting the LoRA rank to 64 and alpha to 128. In the subsequent self-evolution phase, we configure the number of candidate generations to 2 with a sampling temperature of 0.9, and adjust the learning rate to $1 \times 10^{-6}$. The ForeAgent is trained on 8 NVIDIA A100 GPUs, with the training duration set to 2 epochs for the pre-training stage and 1 epoch for the self-evolution stage, respectively. We treat NPR as an external tool (without any parameter updates) and integrate its predictions into the MLLM via tool calling. Following standard protocols in AI-generated content detection, we evaluate our method using classification accuracy for real and fake samples.

\subsection{Experimental Results}
\subsubsection{\textbf{Results On AIGCDetectBenchmark}}
% \paragraph{Results on AIGCDetectBenchmark.}

Table~\ref{tab:aigcdetectbenchmark} summarizes the per-generator accuracy across 16 generators.
Overall, \textbf{ForeAgent} achieves the highest mean accuracy of \textbf{93.3\%}, slightly outperforming the strongest baselines AIGI-Holmes (93.2\%) and AIDE (92.8\%).
Notably, ForeAgent demonstrates clear advantages on modern diffusion-based generators.
For instance, it achieves 96.0\% on Midjourney (vs.\ 81.6\% for AIGI-Holmes and 77.2\% for AIDE), 98.9\% on DALLE2 (vs.\ 85.3\% and 96.6\%), and 98.3\%/98.2\% on Stable Diffusion v1.4/v1.5 (vs.\ 91.3\%/91.4\% and 93.0\%/92.9\%), consistently surpassing both leading baselines by significant margins.
Similarly strong results are observed on GLIDE (97.0\%), Wukong (97.1\%), and VQDM (95.6\%), indicating robust generalization to contemporary AIGC pipelines.
In contrast, open-source MLLM baselines without forensic-specific design (Qwen3-VL-8B/32B) lag behind substantially (54.9\% and 61.0\% mean), confirming that generic visual understanding alone is insufficient for fine-grained AIGC forensics.
We note that ForeAgent is less competitive on WFIR (63.4\%), where several traditional CNN-based detectors (\emph{e.g.}, CNNSpot at 91.7\%, AIDE at 94.2\%) perform considerably better, suggesting that certain sources with distinct post-processing or domain-specific characteristics remain challenging and may benefit from further source-aware augmentation or improved spatial-frequency fusion strategies.

% \paragraph{Results on AIGCDetectBenchmark.}
% Table~\ref{tab:aigcdetectbenchmark} summarizes performance across 16 generators.
% Overall, \textbf{ForeAgent} achieves a mean accuracy of \textbf{93.3\%}, which is competitive with the strongest baseline \textbf{AIGI-Holmes} (93.16\%) and close to \textbf{AIDE} (92.77\%).
% Notably, ForeAgent attains particularly strong results on modern diffusion-based generators (e.g., ADM, GLIDE, Midjourney, Stable Diffusion v1.4/v1.5, VQDM, Wukong, and DALLE2), where it consistently exceeds $95\%$ on most sources, indicating robust generalization to contemporary AIGC pipelines.
% In contrast, open-source MLLM baselines (Qwen3-VL-8B/32B) lag behind substantially (54.88\% and 61.00\% mean), suggesting that generic visual understanding alone is insufficient for fine-grained AIGC forensics.
% We observe that ForeAgent is less competitive on \textbf{WFIR} (57.25\%), highlighting that certain sources with distinct post-processing or data characteristics remain challenging and may require further source-specific augmentation or improved spatial-frequency modeling.

% 实验 Chameleon
%----------------------------------------------------------
\subsubsection{\textbf{Results On Chameleon}}
\label{sec:main_results}
%----------------------------------------------------------
\begin{table}[t]
\centering
\caption{Comparison on Chameleon. We report accuracy(\%) for real, fake, and overall image detection. The best and second-best results are highlighted in \textbf{bold}, \underline{underlined}, respectively.}
\label{tab:chameleon}
\resizebox{\linewidth}{!}{
\begin{tabular}{lccc}
\toprule
\textbf{Method} & \textbf{Real} & \textbf{Fake} & \textbf{Overall} \\
\midrule
\multicolumn{4}{l}{\textit{Feature-based Forensic Methods}} \\
Dire & 99.73 & 2.09 & 57.83 \\
PatchCraft & 96.52 & 1.39 & 55.70 \\
NPR & \textbf{100.00} & 1.68 & 57.81 \\
UnivFD & 41.56 & 85.52 & 60.42 \\
AIDE & 95.06 & 26.80 & 65.77 \\
Effort & 41.66 & 85.92 & 60.65 \\
\midrule
\multicolumn{4}{l}{\textit{MLLM-based Methods}} \\
Qwen3-VL-8B-Instruct & 98.67 & 10.03 & 60.63 \\
Qwen3-VL-32B-Instruct & 96.96 & 17.79 & 62.33 \\
Llama-3.2-11B-Vision-Instruct & 76.23 & 28.15 & 55.59 \\
% \midrule
% \multicolumn{4}{l}{\textit{MLLM-Enhanced Methods}} \\
% Fakeshield  & -- & -- & -- \\
AIGI-Holmes (7B) & -- & -- & 75.90 \\
% \midrule
% \multicolumn{4}{l}{\textit{Proprietary Models}} \\
GPT-5-mini & 94.40 & 36.71 & 69.65 \\
GPT-5 & 92.74 & \underline{67.50} & \underline{81.91} \\
\midrule
\textbf{ForeAgent (Ours)} & \underline{96.69} & \textbf{61.86} & \textbf{82.18} \\
\bottomrule
\end{tabular}
}
\end{table}

% \paragraph{\textbf{Results on Chameleon}}
Table~\ref{tab:chameleon} reports per-class and overall accuracy on Chameleon.
Classical forensic detectors (Dire, PatchCraft, and NPR) achieve near-perfect accuracy on real images ($\geq 96.5\%$) but fail on fake images ($\leq 2.1\%$), indicating limited robustness to highly realistic AI-generated content. UnivFD and effort exhibit the opposite bias, performing substantially better on fake images ($\sim85.5\%$) while sacrificing real-image accuracy ($\sim41.6\%$), which leads to only moderate overall performance ($\sim60.4$--$60.7\%$).

Among MLLM baselines, open-source vision-language models (e.g., Qwen3-VL-8B/32B and Llama-3.2-11B-Vision) generally show strong recognition of real images but struggle to detect fakes, resulting in overall accuracy below $63\%$.
AIDE improves the overall score to $65.77\%$, yet its fake accuracy remains limited ($26.80\%$), suggesting that scaling explainable detection still requires stronger artifact sensitivity and more reliable reasoning.

ForeAgent achieves the best overall accuracy (\textbf{82.18\%}), slightly improving real-image accuracy compared to GPT5 (96.69\% vs.\ 92.74\%) while maintaining competitive fake-image accuracy (61.86\%).
Compared with AIDE, ForeAgent yields a substantial gain in overall accuracy (+16.41\%), demonstrating the effectiveness of combining multi-view forensic cues with hindsight-driven self-refinement.
Notably, ForeAgent approaches GPT5 in fake detection while outperforming it in overall accuracy, indicating improved robustness under class imbalance and reduced bias toward either real or fake classes.

\begin{table}[htbp]
\centering
\caption{The comparison of reasoning quality score (rated 0-10). }
\label{tab:reasoning_quality}
\begin{tabular}{lccccc}
\toprule
\textbf{Model} & \textbf{AVG Score}  \\
\midrule
GPT5-mini  & 4.81 \\
GPT5       & 4.15  \\
ForeAgent  & \textbf{5.35} \\
\bottomrule
\end{tabular}
\end{table}

% 推理质量
\subsubsection{\textbf{Reasoning quality evaluation}}
\begin{figure}[htbp]
\centering
\begin{tikzpicture}
    \begin{axis}[
        ybar,                       % 设置为柱状图
        bar width=15pt,             % 柱子宽度
        enlarge x limits=0.25,      % x轴左右留白
        legend style={
            at={(0.5,1.15)},        % 图例位置（放在图上方）
            anchor=north,
            legend columns=-1,      % 图例横向排列
            draw=none,              % 去掉图例边框
            /tikz/every even column/.append style={column sep=0.5cm}
        },
        ylabel={Percentage (\%)},   % y轴标签
        symbolic x coords={Logic Cons., Evidence Suff., Causal Clear.}, % x轴坐标
        xtick=data,                 % 显示所有数据点的x刻度
        nodes near coords,          % 在柱子上方显示数值
        nodes near coords style={font=\tiny, /pgf/number format/.cd, fixed, precision=1}, % 数值格式
        ymin=0, ymax=90,            % y轴范围
        ymajorgrids=true,           % 显示y轴网格线
        grid style=dashed,
        width=0.95\linewidth,       % 图表宽度
        height=6cm,                 % 图表高度
        % 颜色和样式设置
        cycle list={
            {fill=blue!30!white, draw=blue!70!white},      % GPT5-mini 颜色
            {fill=orange!30!white, draw=orange!70!white},  % GPT5 颜色
            {fill=red!60!white, draw=red!80!black}         % ForeAgent 颜色（强调）
        }
    ]

    % GPT5-mini Data
    \addplot coordinates {(Logic Cons., 64.95) (Evidence Suff., 41.30) (Causal Clear., 65.95)};
    
    % GPT5 Data
    \addplot coordinates {(Logic Cons., 73.50) (Evidence Suff., 23.90) (Causal Clear., 66.20)};
    
    % ForeAgent Data
    \addplot coordinates {(Logic Cons., 75.55) (Evidence Suff., 38.35) (Causal Clear., 76.10)};

    \legend{GPT5-mini, GPT5, ForeAgent}
    \end{axis}
\end{tikzpicture}
\caption{Comparison of reasoning quality metrics. While GPT5-mini excels in evidence and GPT5 in logic, \textbf{ForeAgent} achieves the best balance, leading in Logical Consistency and Causal Clarity.}
\label{fig:reasoning_comparison}
\end{figure}
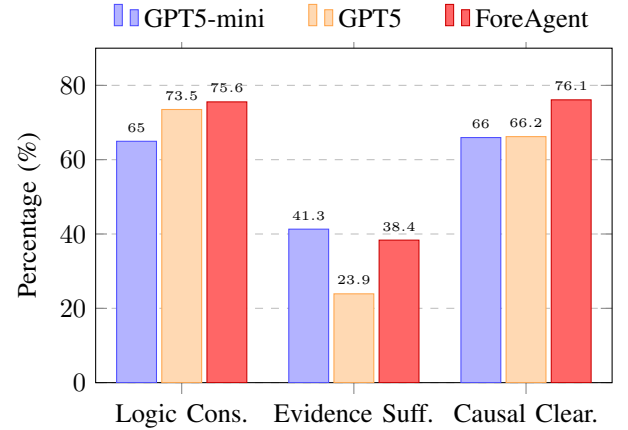

To evaluate the quality of reasoning generated by different models, we randomly sample 2,000 instances from the prediction file generated by GPT5 on AIGCDetectBenchmark, and employ \href{https://blog.google/products-and-platforms/products/gemini/gemini-3-flash/}{Gemini-3-flash-preview} as an external judge. We assess reasoning quality across four dimensions: \textit{quality\_score (Rated 0-10)}, \textit{logic\_consistent} (logical coherence), \textit{evidence\_sufficient} (adequacy of supporting evidence), and \textit{causal\_clear} (clarity of causal reasoning). The evaluation results are summarized in Table~\ref{tab:reasoning_quality}, Figure~\ref{fig:reasoning_comparison}.

Table~\ref{tab:reasoning_quality} presents a comparison of reasoning quality across different models. Notably, GPT5-mini (4.81) surpasses GPT5 (4.15), indicating that model scale alone does not guarantee superior reasoning quality. Our proposed ForeAgent achieves the highest overall score (\textbf{5.35}), outperforming both baseline models by a substantial margin. This improvement demonstrates that ForeAgent's self-evolution paradigm effectively enables adaptive calibration of reasoning depth, leading to more robust and balanced performance.
% The results in Table~\ref{tab:reasoning_quality} reveal distinct reasoning behaviors across models. GPT5-mini achieves an overall score (4.81) surpass GPT5 (4.15) Our proposed ForeAgent effectively combines the strengths of both: it attains the highest overall score (\textbf{5.35}). This indicates that the self-evolution paradigm enables ForeAgent to adaptively calibrate reasoning depth, achieving superior and more balanced performance.
% ForeAgent achieves the highest judged quality under Prompt2 (5.35) and the best average score (4.36), surpassing GPT5 (3.74) and GPT5-mini (4.34).

Figure~\ref{fig:reasoning_comparison} shows that GPT5-mini achieves lower logical consistency (65.0\%), suggesting it excels on \textit{easy} samples with clear artifacts yet struggles with \textit{hard} cases involving subtle cues. Conversely, GPT5 demonstrates stronger logical consistency (73.5\%), reflecting a tendency toward hedged, moderate-quality reasoning that prioritizes stability over decisiveness. ForeAgent benefits from hindsight-driven self-refinement, which attains the strongest average pass rates on \emph{logical consistency} (75.6\%) and \emph{causal clarity} (76.1\%), while maintaining competitive \emph{evidence sufficiency} (38.4\%) compared to GPT5-mini.

\subsection{Ablation study} 
\begin{table}[t]
\centering
\caption{Comparison on Chameleon. We evaluate SFT, ForeAgent without NPR, and the four the discrete wavelet transform: approximation (cA), horizontal detail (cH), vertical detail (cV), and diagonal detail (cD). (\%) }
\label{tab:ablation_wavelet}
\setlength{\tabcolsep}{4pt}
\begin{tabular}{lccc}
\toprule
\textbf{Wavelet Type} & \textbf{Real} & \textbf{Fake} & \textbf{Overall} \\
\midrule
Qwen3-VL-8B-SFT & 54.69 & 77.32 & 64.40 \\
\midrule
+ cA (Approximation) & 50.68 & 88.69 & 66.99 \\
+ cH (Horizontal)    & 31.85 & 91.36 & 57.38 \\
+ cV (Vertical)      & 50.67 & 85.77 & 65.73 \\
+ cD (Diagonal)      & 86.37 & 53.19 & \textbf{72.13} \\
\midrule
ForeAgent wo NPR  & 96.12 & 41.65 & 72.41 \\
ForeAgent (full)     & 95.26 & 43.46 & 73.03 \\
\bottomrule
\end{tabular}
\end{table}

\subsubsection{Effect of Wavelet Decomposition Components and NPR.}
To investigate which sub-band from the discrete wavelet transform (DWT) provides the most discriminative frequency-domain cues for AI-generated image detection, we conduct a systematic comparison across four wavelet components. All experiments use Qwen3-VL-8B with LoRA fine-tuning (rank=64, alpha=128) on 120K training samples from the Genimage. As shown in Table~\ref{tab:ablation_wavelet}, the four sub-bands are defined as follows: 
\textbf{cA (Approximation):} Low-frequency coefficients that preserve the coarse structure and global content of the image. 
\textbf{cH (Horizontal):} Horizontal detail coefficients that capture vertical edges and horizontal texture variations.
\textbf{cV (Vertical):} Vertical detail coefficients that capture horizontal edges and vertical texture variations.
\textbf{cD (Diagonal):} Diagonal detail coefficients that capture diagonal edges and oblique texture patterns.

\noindent\textbf{Sub-band selection.}
cD yields the highest overall accuracy (72.13\%), substantially outperforming cA (66.99\%), cV (65.73\%), and cH (57.38\%). It also achieves the most balanced performance between real and fake classes (86.37\% vs.\ 53.19\%), whereas other sub-bands exhibit pronounced bias toward the fake class. We attribute this to the fact that modern generative pipelines often employ convolutional up-sampling operations that introduce characteristic artifacts—such as checkerboard patterns and irregular edge transitions—predominantly along diagonal orientations, making cD a more informative signal for forensic analysis.
% The results reveal that the \textbf{diagonal detail component (cD)} yields the highest overall accuracy (72.13\%), substantially outperforming cA (66.99\%), cV (65.73\%), and cH (57.38\%). Notably, cD achieves the most balanced performance between real and fake classes (86.37\% vs.\ 53.19\%), whereas other sub-bands exhibit a pronounced bias toward the fake class at the expense of real-image accuracy. We attribute this advantage to the fact that modern generative pipelines (\emph{e.g.}, diffusion models, GANs) often employ convolutional up-sampling operations that introduce characteristic artifacts—such as checkerboard patterns and irregular edge transitions—predominantly along diagonal orientations. These subtle yet systematic anomalies are more salient in the cD sub-band, rendering it a more informative signal for forensic analysis.

\noindent\textbf{Effect of NPR.}
Building upon cD, we further examine the contribution of NPR. As shown in Table~\ref{tab:ablation_wavelet}, incorporating NPR into ForeAgent slightly improves the overall accuracy from 72.41\% to 73.03\%, while also yielding a more balanced trade-off between real-image accuracy (95.26\% vs.\ 96.12\%) and fake-image accuracy (43.46\% vs.\ 41.65\%). This indicates that NPR provides complementary local statistical cues that enhance the model's discriminative capability beyond what frequency-domain features alone can offer.

Based on these findings, we adopt cD combined with NPR as the default frequency-domain representation in ForeAgent.

\subsubsection{Effect of Iterations.}
\begin{figure}[htbp]
\centering
\begin{tikzpicture}
\begin{axis}[
    xlabel={Iteration},
    ylabel={Accuracy (\%)},
    xmin=0, xmax=5,
    ymin=40, ymax=100,
    xtick={0,1,2,3,4,5},
    legend pos=south east,
    grid=major,
    width=0.85\columnwidth,
    height=6cm,
]
\addplot[color=blue, mark=square*, thick] coordinates {(0,73.03)(1,77.78)(2,78.19)(3,79.59)(4,80.10)(5,82.18)};
\addplot[color=green, mark=triangle*, thick] coordinates {(0,95.26)(1,96.92)(2,96.98)(3,95.29)(4,95.30)(5,96.69)};
\addplot[color=red, mark=o, thick] coordinates {(0,43.46)(1,52.32)(2,53.20)(3,58.70)(4,59.88)(5,61.86)};
\legend{Accuracy, Real Acc, Fake Acc}
\end{axis}
\end{tikzpicture}
\caption{Performance trends across iterations on Chameleon.}
\label{fig:iteration_trend}
\end{figure}
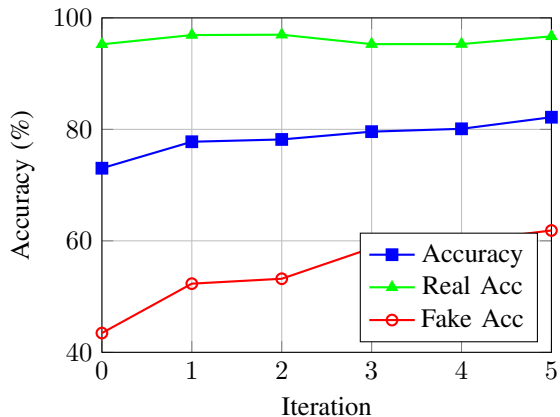

Table~\ref{fig:iteration_trend} illustrates the impact of iteration numbers on model performance. The overall accuracy improves consistently from 73.03\% (iteration 0) to \textbf{82.18\%} (iteration 5), representing a gain of 9.15 percentage points. Notably, while Real Acc remains relatively stable around 95\%-97\% across all iterations, Fake Acc exhibits substantial improvement from 43.46\% to 61.86\%, an increase of 18.40 percentage points. This demonstrates that the self-evolution mechanism primarily enhances the model's ability to detect fake news, which is typically the more challenging task, while maintaining robust performance on real news classification.

% \paragraph{The effect of Quality-Gating model.} \label{ablation:quality_gating} 
% Quality verification was conducted using the single model Qwen3-VL-8B-Instruct.

\subsubsection{The effect of Quality-Gating model.} \label{ablation:quality_gating}
We adopt a dual-expert quality-gating module that combines ForeAgent (Qwen3-VL-8B) and Qwen3-VL-Plus, motivated by three considerations: 1)\textbf{Complementarity}: As shown in Table~\ref{tab:gating_ablation}, the two individual models exhibit complementary strengths—the 8B model yields higher reasoning quality ($4.53\ vs.\ 4.36$) owing to its shared architectural lineage with ForeAgent, which provides a consistent ``internal'' quality perspective on the generated rationales; the more capable Plus model achieves higher detection accuracy ($92.2\% \ vs.\ 91.5\%$) thanks to its stronger visual reasoning capabilities, offering a reliable ``external'' assessment. Their combination leverages both advantages, achieving the best accuracy \textbf{93.3\%} and reasoning quality \textbf{5.12} simultaneously. 2)\textbf{Bias mitigation}: Relying solely on the 8B model to evaluate its own generations may introduce self-reinforcing biases, where the model systematically overlooks its own failure modes. The Plus model serves as an independent cross-validator, breaking this self-evaluation loop and enabling more objective quality assessment. 3)\textbf{Synergistic filtering}: A sample is retained only when \emph{both} experts agree on its quality, enforcing a stricter consensus-based criterion that reduces the risk of low-quality rationales passing through either model's blind spots. The dual-expert strategy improves accuracy by 1.8\%/1.1\% and reasoning quality by 13.0\%/17.4\% over the 8B-only and Plus-only baselines, respectively, confirming that the two experts provide genuinely complementary supervision signals.
% (3) \textbf{Synergistic filtering}: A sample is retained only when \emph{both} experts agree on its quality, enforcing a stricter consensus-based filtering criterion. This joint agreement mechanism reduces the risk of low-quality rationales passing through either model's blind spots, resulting in a cleaner and more reliable training set. The empirical results in Table~\ref{tab:quality_gating} validate this design: the dual-expert strategy improves accuracy by 1.8\% over the 8B-only baseline and 1.1\% over the Plus-only baseline, while boosting reasoning quality by 13.0\% and 17.4\%, respectively. These consistent gains across both metrics confirm that the two experts provide genuinely complementary supervision signals that neither can achieve alone.

\begin{table}[htbp]
\centering
\caption{Comparing different quality-gating strategies on AIGCDetectBenchmark.}
\label{tab:gating_ablation}
\begin{tabular}{lcc}
\toprule
\textbf{Gating Strategy} & \textbf{Accuracy} & \textbf{Reasoning Quality} \\
\midrule
Qwen3-VL-8B Only             & 91.5\% & 4.53 \\
Qwen3-VL-Plus Only           & 92.2\% & 4.36  \\
Dual-Expert (Ours)           & \textbf{93.3\%} & \textbf{5.12} \\
\bottomrule
\end{tabular}
\end{table}

\subsubsection{\textbf{Self-evolution vs. GPT-synthesized Supervision}}
A natural question arises: can we simply leverage powerful LLMs to synthesize reasoning supervision for training? To investigate this, we construct a baseline where GPT5 is provided with images and ground-truth labels to generate reasoning processes, yielding 100K synthetic SFT samples. We then fine-tune the same pre-trained backbone on this dataset for fair comparison.

As shown in Table~\ref{tab:sft_comparison}, while the GPT-synthesized SFT approach achieves a higher Fake Acc (69.62\% vs. 61.86\%), it substantially underperforms on Real Acc (90.07\% vs. 96.69\%) and overall accuracy (81.29\% vs. 82.18\%). More critically, the reasoning quality score is notably lower (4.44 vs. 5.35). These results reveal two key insights: (1) GPT-synthesized reasoning, despite being generated by a more powerful model, may introduce noise or hallucinations that degrade the model's ability to correctly identify authentic content, leading to over-detection of fake news; (2) The self-evolution paradigm produces higher-quality reasoning traces that are better aligned with the model's own capabilities, resulting in more reliable and balanced detection performance. This validates our design choice of adopting self-evolution over direct knowledge distillation from external LLMs.

\begin{table}[htbp]
\centering
\caption{Comparison between GPT-synthesized SFT data and self-evolution paradigm.}
\label{tab:sft_comparison}
\begin{tabular}{p{1.6cm}cccc}
\toprule
\textbf{Method} & \textbf{Real Acc} & \textbf{Fake Acc} & \textbf{Overall} & \textbf{Reasoning Score} \\
\midrule
Qwen3-VL-8B (GPT*) & 90.07\% & 69.62\% & 81.29\% & 4.44 \\
ForeAgent & \textbf{96.69\%} & 61.86\% & \textbf{82.18\%} & \textbf{5.35} \\
\bottomrule
\end{tabular}
\end{table}

\section{Overview of Error Analysis on AIGCDetectBenchmark}

We conduct error analysis on 65\% of the misclassified samples (5,146 cases) from the AIGCDetectBenchmark test set. As shown in Table~\ref{tab:error_distribution}, the error distribution exhibits a significant imbalance: false negatives (Fake$\rightarrow$Real) account for 86.82\% (4,468 cases), vastly outnumbering false positives (Real$\rightarrow$Fake) at 13.18\% (678 cases). This 6.6:1 ratio reveals that the detection system fails primarily by missing fake images rather than by incorrectly flagging real ones, indicating a systematic bias toward classifying images as ``real'', likely due to the high visual quality of modern generative models that successfully deceive the visual reasoning process.

\begin{table}[h]
\centering
\caption{Distribution of Error Types}
\label{tab:error_distribution}
\begin{tabular}{lcc}
\toprule
\textbf{Error Type} & \textbf{Count} & \textbf{Percentage} \\
\midrule
False Negative (Fake$\rightarrow$Real) & 4,468 & 86.82\% \\
False Positive (Real$\rightarrow$Fake) & 678 & 13.18\% \\
\midrule
\textbf{Total} & \textbf{5,146} & \textbf{100\%} \\
\bottomrule
\end{tabular}
\end{table}

\subsection{Round 2 Flip Error Analysis}

\paragraph{Reasoning Consistency Between Rounds.}
We categorize all error samples based on the agreement between Round 1 and Round 2 predictions. Among the 5,146 errors, \textbf{1,450 samples (28.18\%)} represent cases where Round 1 made the correct prediction but Round 2 incorrectly reversed it, while the remaining 3,696 cases (71.82\%) are consistent errors across both rounds. This indicates that the second-round verification mechanism introduces a non-negligible number of new errors rather than purely correcting existing ones.

Further directional analysis of these 1,450 flip errors reveals an extreme asymmetry: 1,431 cases (98.66\%) involve flipping correct ``fake'' predictions to incorrect ``real,'' while only 19 cases (1.34\%) go in the opposite direction. This stark imbalance points to a systematic vulnerability in Round 2 that strongly favors ``real'' classifications. Moreover, when confidence scores increase from Round 1 to Round 2, 93.4\% of such cases are false negatives, suggesting that the model becomes \emph{more} confident in its incorrect ``real'' predictions during the review process, with an overall mean confidence of 0.9617 on misclassified samples.

\paragraph{Reasoning Patterns and Root Causes.}
To understand why Round 2 incorrectly overturns correct predictions, we analyze the reasoning types and justifications employed in flip errors (Table~\ref{tab:flip_reasoning}).

\begin{table}[h]
\centering
\caption{Round 2 Reasoning Patterns in Flip Errors}
\label{tab:flip_reasoning}
\begin{tabular}{lcc}
\toprule
\textbf{Reasoning / Justification} & \textbf{Count} & \textbf{\%} \\
\midrule
\multicolumn{3}{l}{\textit{Analysis Types Used by Round 2}} \\
\quad Texture Analysis & 1,275 & 87.9\% \\
\quad Lighting Analysis & 1,249 & 86.1\% \\
\quad AI Artifact Claims & 1,028 & 70.9\% \\
\quad Anatomical Analysis & 714 & 49.2\% \\
\quad Physical Impossibility & 82 & 5.7\% \\
\midrule
\multicolumn{3}{l}{\textit{Justifications for ``Real'' Classification}} \\
\quad Overall Coherence & 1,189 & 83.1\% \\
\quad Consistent Lighting & 754 & 52.7\% \\
\quad Natural Texture & 428 & 29.9\% \\
\quad Physical Plausibility & 231 & 16.1\% \\
\quad Correct Anatomy & 228 & 15.9\% \\
\bottomrule
\end{tabular}
\end{table}

The results reveal three root causes of Round 2 failures:
\begin{enumerate}
    \item \textbf{Visual Appearance Bias}: Round 2 heavily relies on texture (87.9\%) and lighting (86.1\%) analysis—surface-level features that modern generators excel at synthesizing. The most common justification for overturning to ``real'' is ``overall coherence'' (83.1\%), reflecting a \emph{coherence fallacy}: the assumption that visual consistency implies authenticity.
    \item \textbf{Frequency Domain Distrust}: In 87.3\% of false negative flip cases, Round 2 expresses skepticism toward wavelet analysis, often dismissing frequency-domain evidence as ``technical artifacts'' rather than valid forensic signals. This leads Round 2 to override correct Round 1 judgments that were grounded in spectral cues.
    \item \textbf{Overconfidence}: In the 0.98--1.00 confidence range, 97.7\% of errors are false negatives, indicating severe miscalibration where the model is most confident precisely when it is wrong about ``real'' predictions.
\end{enumerate}

\begin{figure}[t]
\centering
\includegraphics[width=0.4\linewidth]{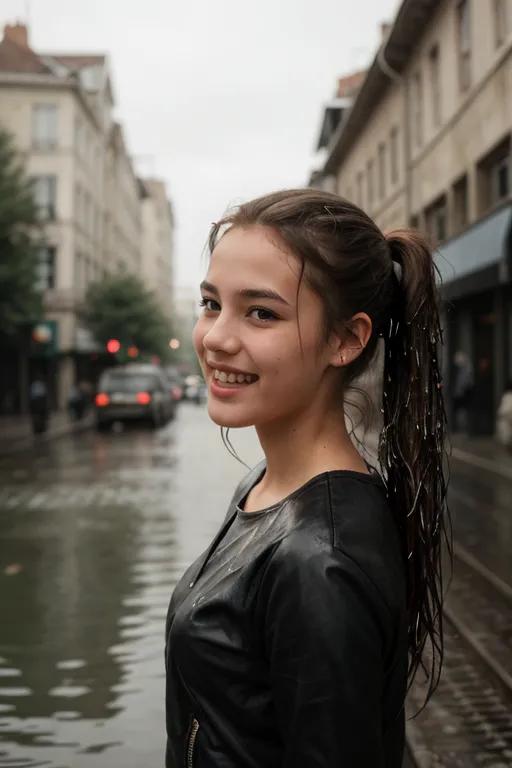}
\vspace{2pt}

\noindent\fcolorbox{black}{green!5}{
\parbox{0.95\linewidth}{\small
\textbf{Perception} \hfill \texttt{Fake | 0.95} \\[2pt]
\textbf{Reasoning} Facial skin exhibits unnaturally uniform 
smoothness without pores; water droplets on the jacket 
appear overly regular in size and distribution. Wavelet coefficients show abnormal 
high-frequency attenuation in facial regions; NPR map 
highlights irregular patterns along hair boundaries. \\[2pt]
\textbf{Result:} \textbf{Fake} \cmark
}}

\vspace{2pt}

\noindent\fcolorbox{black}{red!5}{
\parbox{0.95\linewidth}{\small
\textbf{Verdict} \hfill \texttt{Real | 0.85} \\[2pt]
\textbf{Reasoning} Natural wet-hair rendering, coherent 
rainy-day lighting, and plausible depth-of-field blur. 
The ``smoothness'' is consistent with overcast soft 
lighting. Wavelet patterns are attributable to 
bokeh and JPEG compression, not generative artifacts. \\[2pt]
\textbf{Result:} \textbf{Real} \xmark
}}

\caption{Flip error case. Ground truth: \textbf{Fake}. During the perception round, ForeAgent correctly identifies AI-generation artifacts, but ForeAgent dismisses frequency-domain evidence in favor of visual coherence, incorrectly overriding to ``Real'' in the verdict round.}
\label{fig:case_study_flip}
\end{figure}

\begin{figure}[t]
\centering
\includegraphics[width=0.4\linewidth]{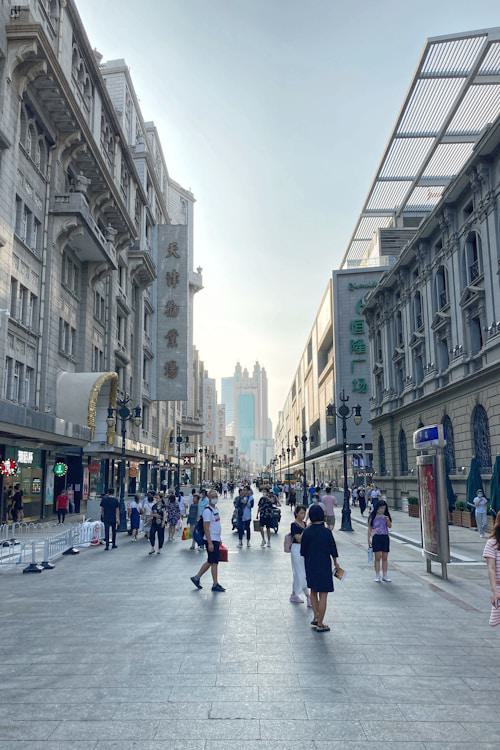}
\vspace{2pt}

\noindent\fcolorbox{black}{red!5}{
\parbox{0.98\linewidth}{\small
\textbf{Perception} \hfill \texttt{Fake | 0.95} \\[2pt]
\textbf{Reasoning} Crowd reveals repetitive postures and 
spatial intervals; façades show misaligned windows; 
Chinese signage displays character-level distortions. Wavelet cD sub-band reveals abnormal 
high-frequency attenuation; NPR map highlights irregularities around pedestrian boundaries. \\[2pt]
\textbf{Result:} \textbf{Fake} \xmark
}}

\vspace{2pt}

\noindent\fcolorbox{black}{green!5}{
\parbox{0.98\linewidth}{\small
\textbf{Verdict} \hfill \texttt{Real | 0.98} \\[2pt]
\textbf{Reasoning} Coherent commercial street with consistent 
perspective, natural crowd behavior, and plausible overcast lighting across architectural elements.  Wavelet patterns appear within normal range, consistent with typical urban photography. \\[2pt]
\textbf{Result:} \textbf{Real} \cmark
}}

\caption{Successful correction case. During the perception round, ForeAgent over-interprets normal photographic characteristics as AI-generation artifacts, while ForeAgent correctly recognizes the scene's authentic visual consistency and re-interprets frequency-domain patterns in the verdict stage.}
\label{fig:case_study_correct}
\vspace{-2pt}
\end{figure}

\paragraph{Case Study.}
We present two representative examples to illustrate both the 
failure and success modes of the dual-round reasoning mechanism. 
In Figure~\ref{fig:case_study_flip}, Round~1 correctly identifies 
a fake portrait by detecting facial over-smoothing and wavelet 
anomalies, yet Round~2 overrides this judgment by attributing 
these artifacts to natural photographic conditions (soft lighting, 
bokeh), exemplifying the \textit{coherence fallacy} and 
\textit{frequency domain distrust} patterns. In contrast, 
Figure~\ref{fig:case_study_correct} demonstrates the intended 
benefit of dual-round verification: Round~1 over-interprets 
normal photographic features of a real street scene as 
AI-generation artifacts (a false alarm), while Round~2 correctly 
recognizes the scene's authentic consistency and rectifies the 
initial misjudgment. Together, these two cases reveal that 
Round~2 succeeds when it leverages \textit{holistic visual 
reasoning} to suppress false positives, but fails when it 
applies the same reasoning to dismiss valid \textit{local 
forensic evidence}, inadvertently introducing false negatives 
that modern generators are specifically optimized to exploit.

\vspace{-2pt}

\section{Related work}
The rapid proliferation of generative models, such as GANs and Diffusion Models, has made the identification of AI-generated images (AIGI) a pivotal challenge in digital forensics. Current research can be broadly categorized into feature-based discriminative models and Large Vision-Language Model (LVLM) based explainable detectors.

\paragraph{Feature-based detection} 
Early methods \cite{8803661,wang2019cnngenerated,fredetect,karras2018progressive,effort,tan2023rethinking,Chameleon} primarily utilized ResNet or ViT backbones to extract statistical artifacts. NPR rethink the generator architectures and identify structural traces left by up-sampling operations, proposing Neighboring Pixel Relationships as a generalizable feature. To address the limitations of single-modality features, AIDE employs a mixture-of-experts approach, combining semantic embeddings from OpenCLIP with low-level frequency patterns derived from DCT scoring and SRM filters. Furthermore, Effort \cite{effort} identifies the "asymmetry phenomenon" where models overfit to monotonous fake patterns; it utilizes Singular Value Decomposition (SVD) to decompose the feature space into orthogonal semantic and forgery subspaces, thereby preserving pre-trained knowledge while enhancing generalization. However, these methods often treat detection as a "black-box" binary classification, lacking logical transparency.

\paragraph{MLLMs-based detection} 
The integration of MLLMs \cite{SIDA,fakereasoning,yang2024researchabilityllmtamperdetection,zhang2024commonsensereasoningdeepfake,zhou2025aigi,kang2025legion} has opened a new frontier for interpretable forensics. AIGI-Holmes couples LLaVA with low-level experts (e.g., NPR) and utilizes a three-stage pipeline (visual pre-training, SFT, and DPO) to provide human-verifiable explanations. Similarly, FakeShield and SIDA extend the task to include tampered region localization and conversational interaction. Regarding reasoning logic, FakeReasoning introduces a structured Chain-of-Thought (CoT) to decompose the reasoning process into summary, caption, reasoning, and conclusion stages. Despite their progress, existing MLLMs-based methods often rely on static fine-tuning datasets and lack a mechanism for iterative self-correction.

\section{Conclusion}
In this paper, we introduced \textbf{ForeAgent}, an agentic forensic framework for AI-generated image detection that addresses two key limitations of existing MLLM-based approaches: insufficient sensitivity to fine-grained artifacts and heavy reliance on proprietary models or costly expert annotations. ForeAgent employs a \textit{Perception--Verdict} architecture to integrate semantic, spatial, and frequency-domain cues, producing coherent and logically grounded decisions via an MLLM-based verdict module. To enable continual self-improvement without human supervision, we proposed a \textit{Hindsight-Driven Self-Refining} mechanism that iteratively reflects on erroneous predictions and applies dual-expert quality gating to curate high-quality fine-tuning samples.Extensive experiments validate the effectiveness of our approach. On the Chameleon benchmark, ForeAgent achieves state-of-the-art accuracy of 82.18\%, surpassing AIDE by 16.41\%. On AIGCDetectBenchmark, it attains 93.3\% mean accuracy across 16 diverse generators. Qualitative evaluations further show that ForeAgent's interpretable reasoning exceeds that of GPT-5, highlighting the value of hindsight-driven self-refinement for both detection and explainability.

\textbf{Limitations and Future Work}. Despite strong overall performance, ForeAgent shows reduced accuracy on certain sources with distinct characteristics (\emph{e.g.}, WFIR), suggesting that source-specific augmentation or improved spatial-frequency modeling may help. The current framework also relies on a fixed set of multi-view features; adaptive or learnable feature selection could further enhance robustness. Moreover, the iterative hindsight-driven refinement paradigm is naturally compatible with reinforcement learning. Future work will explore online RL-based strategies for continuous, real-time adaptation to improve efficiency and generalization. We also plan to extend the framework beyond binary real/fake classification to broader forensic tasks, including fine-grained manipulation localization and multi-source attribution.

\section*{Acknowledgments}
This work was supported in part by the National Natural Science Foundation of China (NSFC) under Grant No. U24A20326 (Regional Joint Fund Project: ``Theory and Methods of Cloud-Edge Collaborative Lightweight Autonomous Intelligent Computing for Content Generation Scenarios,'' 2025--2028).

% \appendices
% appendix.
% You must declare a $\backslash${\tt{section}} before using any $\backslash${\tt{subsection}} or using $\backslash${\tt{label}} ($\backslash${\tt{appendices}} by itself
%  starts a section numbered zero.)}

% \section*{References Section}
% You can use a bibliography generated by BibTeX as a .bbl file.
%  BibTeX documentation can be easily obtained at:
%  http://mirror.ctan.org/biblio/bibtex/contrib/doc/
%  The IEEEtran BibTeX style support page is:
%  http://www.michaelshell.org/tex/ieeetran/bibtex/
 
%  % argument is your BibTeX string definitions and bibliography database(s)
% %\bibliography{IEEEabrv,../bib/paper}
% %
% \section*{Simple References}
% You can manually copy in the resultant .bbl file and set second argument of $\backslash${\tt{begin}} to the number of references
%  (used to reserve space for the reference number labels box).

\bibliographystyle{IEEEtran}
\bibliography{refs}

% \bibliographystyle{IEEEtran}
% \bibitem{ref1}
% {\it{Mathematics Into Type}}. American Mathematical Society. [Online]. Available: https://www.ams.org/arc/styleguide/mit-2.pdf

% \end{thebibliography}

\newpage

% \section*{Biography Section}
 

% \end{IEEEbiography}



\vfill

\end{document}